# A Differential Smoothness-based Compact-Dynamic Graph Convolutional Network for Spatiotemporal Signal Recovery

Pengcheng Gao, Zicheng Gao, and Ye Yuan, *Member, IEEE*

*Abstract*—High quality spatiotemporal signal is vitally important for real application scenarios like energy management, traffic planning and cyber security. Due to the uncontrollable factors like abrupt sensors breakdown or communication fault, the spatiotemporal signal collected by sensors is always incomplete. A dynamic graph convolutional network (DGCN) is effective for processing spatiotemporal signal recovery. However, it adopts a static GCN and a sequence neural network to explore the spatial and temporal patterns, separately. Such a separated two-step processing is loose spatiotemporal, thereby failing to capture the complex inner spatiotemporal correlation. To address this issue, this paper proposes a Compact-Dynamic Graph Convolutional Network (CDGCN) for spatiotemporal signal recovery with the following two-fold ideas: a) leveraging the tensor M-product to build a unified tensor graph convolution framework, which considers both spatial and temporal patterns simultaneously; and b) constructing a differential smoothness-based objective function to reduce the noise interference in spatiotemporal signal, thereby further improve the recovery accuracy. Experiments on real-world spatiotemporal datasets demonstrate that the proposed CDGCN significantly outperforms the state-of-the-art models in terms of recovery accuracy.

## I. INTRODUCTION

Spatiotemporal signals refer to temporal data gathered across a certain spatial area. [1]. As sensor technology progresses, incorporating more extensive spatiotemporal signals becomes increasingly viable. Consequently, these signals are widely applied across diverse domains [2], spanning from environmental and climate research (e.g., predicting wind patterns and forecasting precipitation) to public safety efforts (e.g., predicting criminal activities) [18]. Furthermore, they enable the analysis of human mobility patterns (e.g., monitoring trajectories) and improve the efficiency of intelligent transportation systems (e.g., predicting traffic congestion). [7][13]. These domains commonly highlight the quality of data [78]. Only complete spatiotemporal signals can be effective [19][26]. Unfortunately, Missing data is one of the most common data quality issues [9][28]. Recovering spatiotemporal signals becomes interesting and challenging [61].

Spatiotemporal signals, particularly those collected from outdoor sensors such as meteorological stations, frequently suffer from missing data due to sensor damage, transmission interruptions, power outages, and other related issues. This incomplete data seriously impair the precision of above tasks [15][16][27][50]. Currently, there are relatively few methods which can be applied in spatiotemporal signals recovery [14]. Conventional statistical methods including mean/median imputation, regression imputation, and expectation-maximization (EM), autoregressive integrated moving average (ARIMA) [64]. With the advancements in machine learning, several machine learning-based techniques have been employed to recover spatiotemporal signals. Batista *et al*. [5] proposed a method called K-nearest neighbors (KNN) and Stekhoven *et al*. [17] firstly used random forest (RF) in deal with missing data. At the same time, tensor factorization-based methods [23] are also used to spatiotemporal signals recovery [20][22][25][29]. Furthermore, these methods often overlook the topological structure of spatiotemporal signals, which is a key factor contributing to their suboptimal performance [46][73].

 Thomas *et al*. [35][65] proposed a deep learning method called graph convolutional network (GCN) to handle the data having topological structure [37][57][71][72]. According to this idea, a GCN-based methods [62][63] has been proposed to recover spatiotemporal signals. However, spatiotemporal signals always change over time. This forms what we called dynamic graph which has the feature of high-dimensional [41]. Most of the existing GCN methods are inadequate for handling such dynamic graphs [38][75]. Then, some methods are designed for dynamic graphs [36] is used to recover spatiotemporal signals. However, this kind of model will extract the spatiotemporal features from two separate modules [74] (like GCN and Long Short-Term Memory (LSTM) [70][79][80]). Such two-step process is a loosely coupled spatiotemporal approach, failing to capture the complex internal spatiotemporal correlations.

To address this issue, this paper proposes a Compact-Dynamic Graph Convolutional Network (CDGCN) to capture the complex inner spatiotemporal correlation, which implements a precise spatiotemporal signal recovery. Our contributions can be summarized as follows:

- We establish a CDGCN model to recover spatiotemporal signals accurately. It first leverages the tensor M-product to build a unified tensor graph convolution framework, which considers both spatial and temporal patterns simultaneously. And then, a differential smoothness-based objective function is contrasted to reduce the noise interference in spatiotemporal signal.

➢ P. C. Gao and Y. Yuan are with the College of Computer and Information Science, Southwest University, Chongqing 400715, China (e-mail: gao001109@outlook.com, yuanyekl@swu.edu.cn).
➢ Z. C. Gao is with the Hanhong College, Southwest University, Chongqing 400715, China (e-mail: gzcswu@sina.com).

- We collect a complete real-world dataset which includes 127 weather stations from 7 states near the Great Lakes region of the USA, 5 meteorological features and 744 time-steps.

- Empirical experiments demonstrate that CDGCN we designed has good performance in meteorological spatiotemporal signals recovery.

In following sections, section II is the preliminaries, section III gives the CDGCN model, Section IV provides detailed empirical studies and analyses and section V presents the conclusions in the last.

## II. PRELIMINARIES

### A. Spatiotemporal Signals

Spatiotemporal signals with graph structure usually can be divided into three types [45]. The three types can be described by following formal definitions:

***Definition 1:*** Dynamic graph with temporal signal. This kind of spatiotemporal signals is the ordered set of graph and node feature matrix tuples $S=\{(G_1,X_1),…,(G_T,X_T)\}$ where the edges between the vertexes satisfy that $E_t=E$, $t\in\{1,…,T\}$ and $G_t=(V,E_t)$. The node feature matrices are that $X_t\in \mathbf{R}^{N\times F}$, $t\in\{1,…,T\}$ [51].

***Definition 2***: Dynamic graph with static signal. This kind of spatiotemporal signals is the ordered set of graph and node feature matrix tuples $S=\{(G_1,X),…,(G_T,X)\}$ where the edges between the vertexes satisfy that $E_t=E$, $t\in\{1,…,T\}$ and $G_t=(V,E_t)$. The node feature matrices are that $X\in\mathbf{R}^{N\times F}$ [51].

***Definition 3***: Static graph with temporal signal. This kind of spatiotemporal signals is the ordered set of graph and node feature matrix tuples $S=\{(G,X_1),…,(G,X_T)\}$ where the node feature matrices are $X_t\in\mathbf{R}^{N\times F}$, $t\in\{1,…,T\}$ [51].

### B. Graph Convolutional Network

According the GCN model proposed by Thomas *et al*. [35]. GCN can learns the low-dimensional embedding for each node [52][66]. A multi-layer GCN with the following layer-wise propagation rule:

$$H^{(l+1)} = \sigma(AH^{(l)}U^{(l)}), \quad (1)$$

where $H^{(0)}=X$, $X$ is the node feature matrices. $H^{(l)}$ represents the input node embedding, $H^{(l+1)}$ represents the output node embeddings, respectively. $\tilde{A}$ is $A$ which has been normalized. $U^{(l)}$ is the learnable weight matrix of l-th layer. $\sigma(\cdot)$ is the activation function.

### C. Tensor M-Product Framework

A tensor is a multidimensional array. Nth-order tensor is an element of the tensor product of N vector spaces, each of which has its own coordinate system [10][30] [32-34][47][49][88]. A first-order tensor is a vector, a second-order tensor is a matrix, and tensors of order three or higher are called higher-order tensors [87]. In this paper, tensor is third-order tensor and use $W\in\mathbf{R}^{I\times J\times K}$ to denote it. $W_{ij:}$ and $W_{::k}$ denote the tubes and frontal slides of $W$. Based on this theory, we can describe spatiotemporal signals as a third-order tensor $W\in\mathbf{R}^{T\times N\times F}$, T denotes the time series, N denotes the number of nodes, F denotes the number of node features in spatiotemporal signals, respectively.

With the theory of tensor, we will introduce the M-product [38] framework. The most important feature of this framework is that the result of M-product of two third-order tensors is still third-order tensor [53]. Following three parts can describe M-product framework.

- ***M-transform:*** $M\in\mathbf{R}^{T\times T}$ is a mixing matrix. The M-transform of $W\in\mathbf{R}^{T\times N\times F}$ can be represented as $W\times_3 M\in\mathbf{R}^{T\times N\times F}$ and defined elementwise as:

$$\left(W\times_3 M\right)_{tnf} \stackrel{\text{def}}{=} \sum_{k}^{T} M_{tk}W_{tnf}, \quad (2)$$

in above formula, $W\times_3 M$ exists in the transformed space. It is important to note that if $M$ is invertible, after that $(W\times_3 M)\times_3 M^{-1}=M$ [38].

- ***Facewise product:*** If we have two tensors, $W\in\mathbf{R}^{I\times J\times K}$ and $Y\in\mathbf{R}^{J\times Q\times K}$. The facewise product can be defined as $W\Delta Y\in\mathbf{R}^{I\times Q\times K}$ [38]. In other words, the facewise of two tensors which have the same shape of $W$ and $Y$ can be described as $(W\Delta Y)_{::k}= W_{::k}Y_{::k}$.

- ***M-product:*** With the M-transform and facewise product we can describe M-product as following:

$$W*Y \stackrel{\text{def}}{=} \left(\left(W\times_3 Y\right)\Delta\left(Y\times_3 M\right)\right)\times_3 M^{-1}, \quad (3)$$

The initial M-product entails utilizing the Discrete Fourier Transform (DFT) matrix for M [53][54]. The DFT matrix can enhance computational efficiency through the Fast Fourier Transform (FFT).

## III. MODEL

In this section, we will give the CDGCN model which applicates in meteorological spatiotemporal signals recovery. Firstly, the spatiotemporal signals are the static graph with temporal signals which have the feature of high-dimensional [39][40][68] and incomplete [69][76][77], as the presentation in section II. Its framework is exemplified in Fig. 1.

### A. Graph Construction

The meteorological spatiotemporal signals usually collected from ground sensors in weather stations. We can use the geographic location of weather stations to construct graph. We compute the geographic distance matrix D based on the latitude, longitude by Haversine formula [55].

$$D_{ij} = 2R \arcsin\left(\sqrt{\text{hav}(la_j - la_i) + \cos(la_i)\cos(la_j)\text{hav}(lo_j - lo_i)}\right), \quad (4)$$

$R$ is the earth's radius, $hav(a)=sin^2(a/2)$ is the Haversine function. $la$ and $lo$ denotes the latitude and longitude respectively and the subscripts of them represent different weather stations. This is the shortest distance along a sphere's surface, also known as geodesic distance [56]. After that we compute the similarity matrix $P$ between weather stations by Gaussian kernels.

$$P_{ij} = \exp\left(-\frac{D_{ij}^2}{2\theta^2}\right), \quad (5)$$

$D_{ij}$ represents the geographic distance of two nodes and $\theta$ is the hyperparameter which is set as 200 according to dataset. At last, an edge is formed between nodes $i$ and $j$. If $P_{ij}>\omega$ and the weighted adjacency matrix $A$ is given by $P_{ij}$ and $\omega$.

$$A_{ij} = \begin{cases} P_{ij} & \text{if } P_{ij} > \omega \\ 0 & \text{otherwise} \end{cases}, \quad (6)$$

where the weights in $A$ represent the geographic similarity between nodes in graph. We do not perform symmetric normalization on the adjacency matrix computing by Gaussian kernels due to the adjacency matrix already have the symmetry and the weights in it are in the same scale. This adjacency matrix resembles the symmetric normalization adjacency matrix, which addresses the issue of gradient explosion. What's more, it enables the model to leverage data patterns from neighboring stations that share similarities, thereby enhancing its ability to accurately predict missing values for the current site. The validity will be proved in section IV.

### B. Tensor Graph Convolution

The tensor graph convolutional module is the TM-GCN [38] which is inspired by the first order GCN [35] for static graphs. We can extract the graph information including adjacency matrix $A$ and feature matrices $X$ from spatiotemporal signals. The adjacency matrix is given by equations (4) - (6), and the feature matrices are collected by sensors, which are usually incomplete.

The choose of the matrix $M$ which defines the M-product is the most important things for TM-GCN. For temporal signals, $M$ is defined as a lower triangular, which is given by following formula:

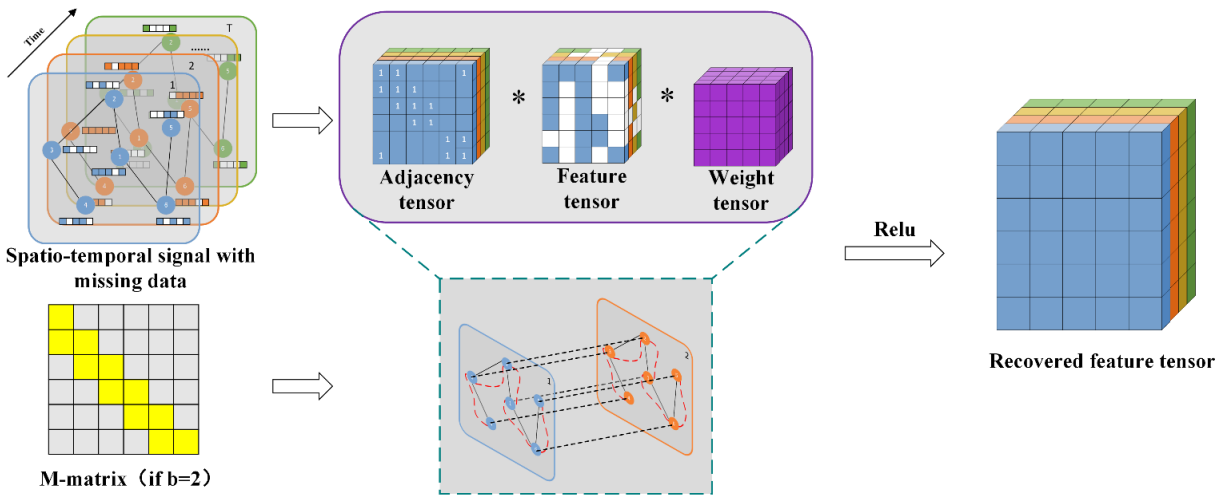

Fig. 1. The framework of CDGCN.

$$M_{tk} \stackrel{\text{def}}{=} \begin{cases} \dfrac{1}{\min(b,t)} & \text{if } \max(1, t-b+1) \leq k \leq t \\ 0 & \text{otherwise} \end{cases}, \quad (7)$$

where $b$ is the bandwidth of $M$, which ensures that the tensor graph convolution module only uses the $b$ time information closest to the current time.

The tensor graph convolution transforms GCN into a tensor model by leveraging the M-product framework. Let $A \in \mathbf{R}^{N \times N \times T}$ be a tensor with frontal slices $A_{:::t}=A$, $A$ is the weighted adjacency matrix. Then, let $W \in \mathbf{R}^{N \times F \times T}$ be a tensor with frontal slices $W_{:::t}=X^{(t)}$, $X^{(t)}$ denotes the incomplete temporal feature matrices. Finally, $U \in \mathbf{R}^{F \times F' \times T}$ is the parameters tensor. We can represent tensor graph convolution as following:

$$H = \sigma(A * W * U), \quad (8)$$

$H$ is the output of the tensor graph convolution module which represents the recovered spatiotemporal signals. $\sigma(\cdot)$ is a nonlinear activation function which we choose ReLU here [4][84-86].

*C. Differential Smoothness-based Objective Function*

We observe that the majority of features in meteorological spatiotemporal signals do not change a lot within 1 hour interval (for instance, temperature may only rise 1 degree from 8:00 AM to 9:00 AM). Based on this characteristic, we have devised a novel approach called differential smoothness regularization to aid in capturing the temporal patterns [8] of these signals effectively. However, it's worth noting that weather features may undergo rapid changes within certain hours (for example, wind speed may transition from 1 to 10 within 1 hour). Yet, such drastic changes tend to occur at specific times, followed by a return to stability (for example, wind speed may subsequently only change from 10 to 11 or remain constant). Even in these scenarios, the proposed differential smoothness regularization can still play a crucial role. The following formula represents the differential smoothness regularization:

$$Reg = \lambda \frac{1}{(T-1) \times N \times F} \sum_{t=1}^{T-1} \sum_{i=0}^{N-1} \sum_{j=0}^{F-1} \left| \hat{y}_{t,i,j} - \hat{y}_{t-1,i,j} \right|, \quad (9)$$

where $\lambda$ is a hyperparameter which control the strength of regularization.

We choose the Huber loss as our loss function. Compared to Mean Absolute Error (MAE) and Mean Squared Error (MSE), the Huber loss function can effectively enhance the robustness of the squared error loss function to outliers [42]. This feature will be helpful when model recovery the meteorological spatiotemporal signals which usually contain outliers [59]. Huber loss has a hyperparameter $\delta$ which control whether choose RSE or MAE. The objective function consists of differential smoothness regularization and Huber loss:

$$L = \text{Loss}(y, \hat{y}) + \lambda \frac{1}{(T-1) \times N \times F} \sum_{t=1}^{T-1} \sum_{i=0}^{N-1} \sum_{j=0}^{F-1} \left| \hat{y}_{t,i,j} - \hat{y}_{t-1,i,j} \right|,$$

$$\text{where Loss}(y, \hat{y}) = \begin{cases} \dfrac{1}{2}(y-\hat{y})^2, & \text{if } |y-\hat{y}| \leq \delta \\ \delta(|y-\hat{y}| - \dfrac{1}{2}\delta), & \text{otherwise} \end{cases} \quad (10)$$

## IV. EMPIRICAL STUDIES

*A. General Settings*

**Evaluation Protocol.** We use the model to recover the spatiotemporal signals. Based on this task, we choose the estimation accuracy as our evaluation protocol. Commonly, the root mean squared error (RMSE) is adopted to measure a model's estimation accuracy [3][12][81-83]. Except that, we choose RSE [43] as another evaluation protocol which can estimate tensors recovery's accuracy:

$$RSE = \frac{\left\| W - W_{origin} \right\|_F}{\left\| W_{origin} \right\|_F}$$

$$RMSE = \sqrt{\frac{1}{(T-1) \times N \times F} \sum_{t=1}^{T-1} \sum_{i=0}^{N-1} \sum_{j=0}^{F-1} (y_{t,i,j} - \hat{y}_{t-1,i,j})^2}$$

where $\|\cdot\|_F$ is the F-norm, $W_{origin}$ represents the complete spatiotemporal signals and $W$ denotes the recovered incomplete spatiotemporal signals.

**Dataset.** We collected the real-word dataset from Great Lakes region of the USA, which includes 127 weather stations and 5 meteorological features (temperature, humidity, pressure, visibility, wind speed) for one month. We show the distributions of the weather stations in Fig. 2. To assess the model's effectiveness, we randomly introduce missing data ranging from 99% to 20% across the dataset. The dataset is then divided into training, validation, and test sets in a ratio of 60%, 20%, and 20% respectively

**Baselines.** To verify the effectiveness of the model [24], we choose 5 other models including ARIMA [64], RF [53], CP decomposition (CP) [22], GCN [35] and EvolveGCN [31]. ARIMA is a statistical method, RF is a machine learning method, CP decomposition is a tensor decomposition method. GCN and EvolveGCN are GNN-based methods which is used to deal with static and dynamic graphs, respectively. All these methods are used to recovery missing data.

TABLE I: THE COMPARISON RESULTS ON ESTIMATION ERRORS.

| Methods | Error | 20% | 45% | 65% | 80% | 90% | 95% | 99% |
|---|---|---|---|---|---|---|---|---|
| ARMIA | RMSE | 3.7685 | 5.6631 | 6.8128 | 7.5613 | 8.0112 | 8.2364 | 8.0112 |
|  | RSE | 6.0688 | 9.1198 | 10.9713 | 12.1768 | 12.9013 | 13.2639 | 12.9013 |
| CP | RMSE | 0.3836 | 0.5629 | 0.6843 | 0.7566 | 0.8006 | 0.8251 | 0.8428 |
|  | RSE | 0.6178 | 0.9065 | 1.1019 | 1.2185 | 1.2894 | 1.3287 | 1.3573 |
| RF | RMSE | 0.0918 | 0.1413 | 0.1684 | 0.1974 | 0.2091 | 0.2268 | 0.2289 |
|  | RSE | 0.1478 | 0.2275 | 0.2713 | 0.3179 | 0.3367 | 0.3652 | 0.3686 |
| GCN | RMSE | 0.0908 | 0.1362 | 0.1641 | 0.1828 | 0.1939 | 0.1992 | 0.2036 |
|  | RSE | 0.1462 | 0.2193 | 0.2643 | 0.2943 | 0.3122 | 0. 3207 | 0.3278 |
| EvolveGCN | RMSE | 0.0904 | 0.1410 | 0.1676 | 0.1854 | 0.1974 | 0.2034 | 0.2049 |
|  | RSE | 0.1456 | 0.2271 | 0.2699 | 0.2985 | 0.3179 | 0.3271 | 0.3300 |
| **CDGCN** | RMSE | **0.0820** | **0.1228** | **0.1484** | **0.1653** | **0.1755** | **0.1802** | **0.1844** |
|  | RSE | **0.1320** | **0.1978** | **0.2390** | **0.2662** | **0.2826** | **0.2902** | **0.2970** |

The percentages in the table column head (for example, 20%) represent the missing ratio in the spatiotemporal signals.

**Hyperparameters.** In the CDGCN we have 5 main hyperparameters including learning rate, weight decay, b, d and l. We set learning rate as {0.05,0.1,0.15} to fit different missing ratios. The order of magnitude of weight decay is 10-3. In addition, b is set as 168, indicating that CDGCN considers 168 time-steps (a week). Depending on various missing ratios, $\lambda$ is chosen from {0.05, 0.15, 0.2, 0.25, 0.35}, which controls the strength of differential smoothness regularization. Finally, $\delta$ is set to 1 which is the default value. The hyperparameters of other models are the values at which the model performs best under different missing rates.

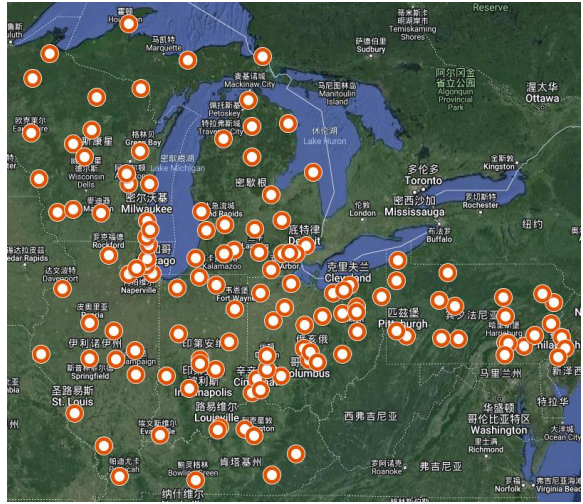

Fig. 2. The Distribution of Weather Stations.

*B. Comparison Performance and Analysis*

We compared CDGCN with other models across various missing ratios of spatiotemporal signals, as shown in Table I and Fig. 3. Lower RMSE and RSE values indicate better model performance. It is evident from Table II and Fig. 3 that CDGCN consistently exhibits the best performance, irrespective of the missing ratio being high or low.

Furthermore, it's apparent that all models achieve the lowest RMSE and RSE values when the missing ratio is 20%.

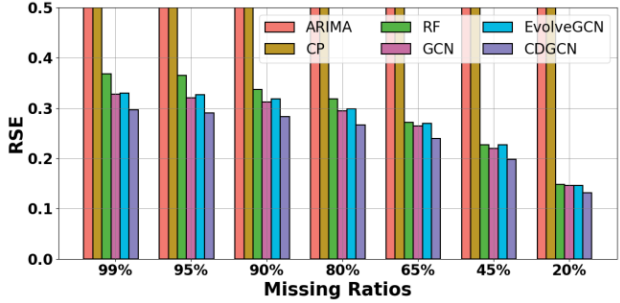
(a) Comparison results on RSE

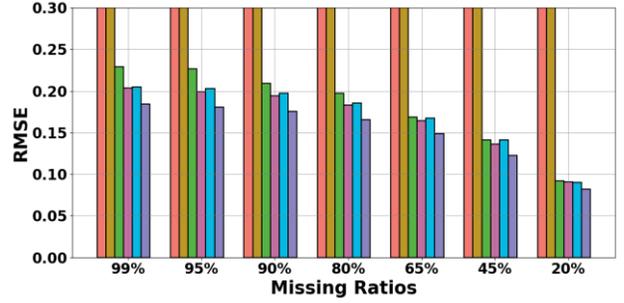
(b) Comparison results on RMSE

Fig. 3. RSE/RMSE of compared models; all panels share (a)'s legend.

TABLE II: ABLATION STUDY.

| Methods | Error | 20% | 45% | 65% | 80% | 90% | 95% | 99% |
|---|---|---|---|---|---|---|---|---|
| **CDGCN** | RMSE | **0.0820** | **0.1228** | **0.1484** | **0.1653** | **0.1755** | **0.1802** | **0.1844** |
|  | RSE | **0.1320** | **0.1978** | **0.2390** | **0.2662** | **0.2826** | **0.2902** | **0.2970** |
| CDGCN | RMSE | 0.0833 | 0.1260 | 0.1514 | 0.1678 | 0.1780 | 0.1833 | 0.1870 |
| (w/o regularization) | RSE | 0.1342 | 0.2029 | 0.2437 | 0.2701 | 0.2866 | 0.2952 | 0.3011 |
| TM-GCN | RMSE | 0.0911 | 0.1322 | 0.1558 | 0.1711 | 0.1816 | 0.1867 | 0.1908 |
| (standard) | RSE | 0.1468 | 0.2128 | 0.2508 | 0.2756 | 0.2924 | 0.3007 | 0.3072 |

The percentages in the table column head (for example, 20%) represent the missing ratio in the spatiotemporal signals.

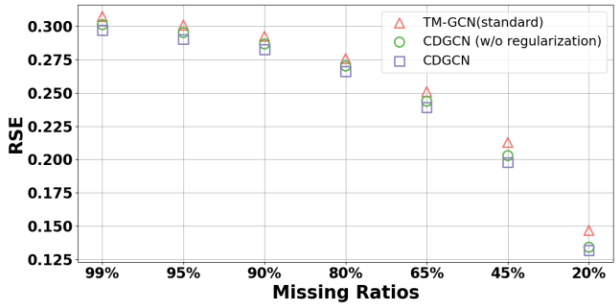
(a) Comparison of RSE on Ablation study

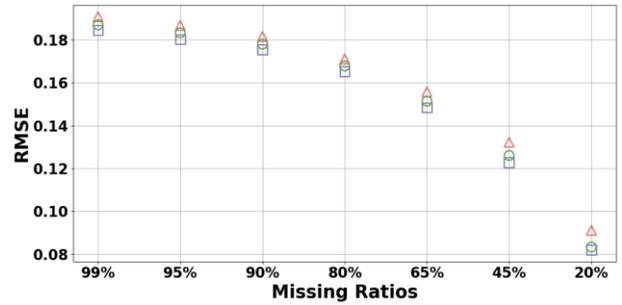
(b) Comparison of RMSE on Ablation study

Fig. 4. RSE/RMSE of Ablation study; all panels share (a)'s legend.

For instance, the RSE values for the six models are 6.0668, 0.6178, 0.1478, 0.1462, 0.1456, and 0.1320, respectively. This indicates that all models are quite effective when the missing ratio is low. However, as the missing ratio increases, GCN-based methods begin to outperform others. The reason behind is that GCN-based methods utilize the topological structure of spatiotemporal signals in recovering missing values. When the missing ratio is 99%, the RSE values for CDGCN and GCN are 0.2970 and 0.3278, respectively. CDGCN significantly outperforms the GCN model. This further emphasizes that GCN, originally designed for static graph processing, is not suitable for handling dynamic graphs like spatiotemporal signals. Additionally, when spatiotemporal signals are highly incomplete, the lack of information poses a significant challenge for models to capture the intrinsic relationships between information. As the missing ratio rises from 20% to 99%, the RMSE of CDGCN is getting higher from 0.0820 to 0.1844. We can draw the conclusion that higher missing ratios not only does the amount of available data decrease, but the structure and patterns within the data become increasingly obscured. This makes it harder for models to accurately impute missing values and effectively reconstruct the spatiotemporal signals. Consequently, the declining performance of all models underscores the intricate nature of dealing with highly incomplete data.

*C. Ablation Study*

The ablation study is used to verify the validity of the differential smoothness regularization and the Gaussian adjacency matrix. We firstly benchmark TM-GCN as the standard model, without regularization and employing symmetric normalization adjacency matrix. Then, we substitute the symmetric normalization adjacency matrix with a Gaussian adjacency matrix. The results are depicted in Table II and Fig. 4. It is evident that the inclusion of differential smoothness regularization aids in improving the model's fit to the dataset. Moreover, compared to the symmetric normalization adjacency matrix, the Gaussian adjacency matrix serves as a more effective representation of the geodesic similarity between weather stations.

*D. Summary*

In accordance with the numerical results and analyses mentioned above, the following perspectives are summarized:

- Compared to ARIMA, RF, and CP, the GCN-based model is better suited for datasets with graph structures, such as spatiotemporal signals. CDGCN, designed for dynamic graphs, performs better in processing spatiotemporal signals.
- The incorporation of differential smoothness regularization improves the model's effectiveness by approximating adjacent time data.
- Compared to the symmetric normalization adjacency matrix, the Gaussian adjacency matrix provides a better representation of the geodesic similarity between different weather stations, aiding in the recovery of spatiotemporal signals.

## V. Conclusion

In this paper, we have proposed a model based on tensor graph neural networks to address the problem of spatiotemporal signal recovery, considering these signals as tensors with a graph structure. Our dataset comprises meteorological spatiotemporal signals. Meteorological data collected from sensors often contain missing values, making it crucial to accurately handle these gaps, given the significance of meteorological information across various research domains [58]. Our proposed model demonstrates excellent performance in recovering meteorological spatiotemporal signals compared to traditional methods [48].

Regarding our forthcoming study, we would like to utilize the recovered spatiotemporal signals to forecast power outages [6]. The increasing dependence of economic activities on electrical power has led scientists and engineers to prioritize enhancing the efficiency and reliability of power grids [44]. Additionally, numerous power outages are attributable to adverse weather conditions affecting the distribution infrastructure [60]. Given this, the complete meteorological information recovered by CDGCN holds particular importance for subsequent prediction tasks and further industrial applications [21][49][67].